\documentclass[runningheads]{llncs}
\usepackage[width=122mm,left=12mm,paperwidth=146mm,height=193mm,top=12mm,paperheight=217mm]{geometry}
\usepackage{graphicx, float, bbding, amsmath, multirow, amssymb}
\usepackage{url}

\begin{document}
\title{TRIG: Transformer-Based Text Recognizer with Initial Embedding Guidance}
%
%
\author{Yue Tao \and
Zhiwei Jia \and
Runze Ma \and Shugong Xu\inst{ (}\Envelope\inst{)}}

\authorrunning{Tao et al.}
%
\institute{School of Communication and Information Engineering, Shanghai University, Shanghai, 200444, China\\
\email{\{yue\_tao, zhiwei.jia, runzema, shugong\}@shu.edu.cn}}

\maketitle              
%


\begin{abstract}
Scene text recognition (STR) is an important bridge between images and text, attracting abundant research attention. While convolutional neural networks (CNNS) have achieved remarkable progress in this task, most of the existing works need an extra module (context modeling module) to help CNN to capture global dependencies to solve the inductive bias and strengthen the relationship between text features. Recently, the transformer has been proposed as a promising network for global context modeling by self-attention mechanism, but one of the main short-comings, when applied to recognition, is the efficiency. We propose a 1-D split to address the challenges of complexity and replace the CNN with the transformer encoder to reduce the need for a context modeling module. Furthermore, recent methods use a frozen initial embedding to guide the decoder to decode the features to text, leading to a loss of accuracy. We propose to use a learnable initial embedding learned from the transformer encoder to make it adaptive to different input images. Above all, we introduce a novel architecture for text recognition, named \textbf{TR}ansformer-based text recognizer with \textbf{I}nitial embedding \textbf{G}uidance (\textbf{TRIG}), composed of three stages (transformation, feature extraction, and prediction). Extensive experiments show that our approach can achieve state-of-the-art on text recognition benchmarks. 

\end{abstract}

\section{Introduction}
STR, aiming to read the text in natural scenes, is an important and active research field in computer vision~\cite{long2021scene,zhu2016scene_survey}. Text reading can obtain semantic information from images, playing a significant role in a variety of vision tasks, such as image retrieval, key information extraction, and~document visual question~answering. 

Among the feature extraction module of existing text recognizers, convolutional architectures remain dominant. For~example, ASTER~\cite{2018ASTER} uses ResNet~\cite{2016Resnet} and SRN~\cite{yu2020srn} uses FPN~\cite{lin2017fpn} to aggregate hierarchical feature maps from ResNet50. As~we all know, the~text has linguistic information and almost every character has a relationship with each other. So features with global contextual information can decode more accurate characters. Unfortunately, the~convolutional neural network (CNN) has an inductive bias on locality for the design of the kernel. It lacks the ability to model long-range dependencies, hence text recognizers should use context modeling structures to gain better performance. It is a common practice that Bi-directional LSTM (BiLSTM) \cite{lstm} is effective to enhance context modeling. Such context modeling modules introduce additional complexity and operations. So a question comes: Why not replace CNN with another network which can model long-range dependencies in a feature extractor without an additional context modeling module?

With the introduction of the transformer~\cite{vaswani2017transformer}, the~question has an answer. Recently, it has been proposed to regard an image as a sequence of patches and aggregate feature in global context by self-attention mechanisms~\cite{2020ViT}. Therefore, we propose to use a pure transformer encoder as the feature extractor instead of CNN. Due to the ability of dynamic attention, global context, and~better generalization of the transformer, the transformer encoder can provide global and robust features without the extra context modeling module. By~this way, we can simplify the four-stage STR framework (transformation stage, feature extraction stage, context modeling stage, and~prediction stage) proposed in Baek et al. \cite{baek2019benchmark} to three stages by removing the need for a context modeling module. Our extensive experiments prove the effectiveness of the three-stage architecture. It shows that the additional context modeling module degrades performance rather than any gain and the feature extractor exactly models long-range dependencies when using the transformer~encoder. 

\begin{figure}[ht]
\centering
\includegraphics[width=12 cm]{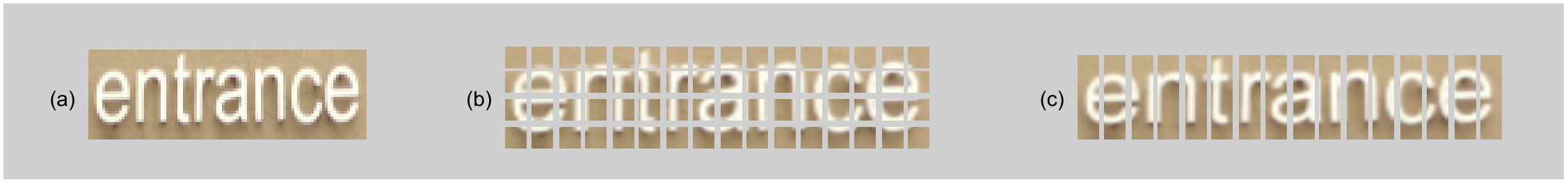}        
\caption{{Difference between 1 and D split and 2-D split.} (\textbf{a}) A rectified text image. (\textbf{b}) 2-D split: The text image is split into square patches proposed in {the} ViT-like backbone. (\textbf{c}) 1-D split: The text image is split into rectangle patches {whose height is the same as the height of the rectified image}. Using the same patch dimension, the~1-D split is more efficient than the 2-D~split.\label{fig_patch}}
\end{figure}

Despite the strong ability of the transformer, the~high demand for memory and computation resources may cause difficulty in the training and inference process. For~example, the~authors of Vision Transformer (ViT) \cite{2020ViT} used extensive computing resources to train their model (about 230 TPUv3-core-days for the ViT-L/16). It is hard for researchers to access such huge resources. The~main reason for the high complexity of the transformer is the self-attention mechanism inside. Complexity and sequence length are squared. Therefore, reducing the sequence length can effectively reduce the complexity. With~the consideration of efficiency, we do not simply use the square patch size proposed in the ViT-like backbone used in image classification, segmentation, and~object detection~\cite{2020deit,beal2020vit-frcnn,zhao2020point,valanarasu2021medicaltransformer}. Instead, we propose the 1-D split to split the picture into rectangle patches whose height is the same as the input image, shown in Figure~\ref{fig_patch}c. In~this way, the~image can convert to a sequence of patches (1-D split) whose length is shorter than the 2-D split (the height of patch size is smaller than the input image). The~design of patch size has the advantage of fewer Multiply Accumulate operations (MACs), which leads to faster training and inference with fewer~resources. 

The prediction stage is another important part of the text recognizer, which decodes the feature to text. An attention-based sequence decoder is commonly used in previous works and has a hidden state embedding to guide the decoder. Recent methods~\cite{2018ASTER,li2019sar,2019Master} use the frozen zero embedding to initialize the hidden state, which remains the same when different images are inputted, influencing the accuracy of the decoder. To~make the hidden state of the decoder adaptive to different inputs, we propose a learnable initial embedding learned from the transformer encoder to dynamically learn information from images. The~adaptive initial embedding can guide the decoding process to reach better~accuracy.

To sum up, this paper presents three main contributions:

\begin{enumerate}
\item[1.] We propose a novel three-stage architecture for text recognition, TRIG, namely \textbf{TR}ansformer-based text recognizer with \textbf{I}nitial embedding \textbf{G}uidance. TRIG leverages the transformer encoder to extract global context feature{s} without an additional context modeling module used in CNN-based text recognizers. Extensive experiments on several public scene text benchmarks demonstrate the proposed framework can achieve state-of-the-art (SOTA) performance.

\item[2.] A~1-D split is designed to divide the text image as a sequence of rectangle patches with the consideration of~efficiency. 

\item[3.] We propose a learnable initial embedding to dynamically learn information from the whole image, which can be adaptive to different input images and precisely guide the decoding~process.
\end{enumerate}

\section{Related~Work}

Most traditional scene text recognition methods~\cite{sw_text1,svtp,sw_text3,sw_text4,sw_text5} adopt a bottom-up approach, which first detects individual characters with a sliding window and classifies them by using hand-crafted features. With~the development of deep learning, top-down methods were proposed. These approaches can be roughly divided into two categories by applying a transformer or not, namely transformer-free methods and transformer-based~methods. 

\subsection{Transformer-Free~Methods}
Before the proposal of the transformer, STR methods only use CNN and recurrent neural network (RNN) to read the text. CRNN~\cite{shi2016crnn} extracts feature sequences using CNN, and~then encodes the sequence by RNN. Finally, Connectionist Temporal Classification (CTC) \cite{graves2006ctc} decodes the sequence to the text results. By~design, this method is hard to address curve or rotated text. To~deal with it, Aster proposes the method of spatial transformer networks (STN) \cite{jaderberg2015stn} with the 1-D attention decoder. Without~spatial transformation~\cite{wan20192d-ctc,li2019sar}, propose methods to handle irregular text recognition by 2-D CTC decoder or 2-D attention decoder. Furthermore, segmentation-based methods~\cite{masktextspotter} can also be used to read text, which should be supervised by character-level annotations. SEED~\cite{qiao2020seed} use{s} semantic information which is supervised by a pre-trained language model to guide the attention~decoder. 

\subsection{Transformer-Based~Methods}
The transformer, first applied to the field of machine translation and natural language processing, is a type of neural network mainly based on the self-attention mechanism. Inspired by NLP success, ViT applies a pure transformer to tackle the image classification tasks and attains comparable results. Then, Data-efficient Image Transformers (DeiT) \cite{2020deit} achieves competitive accuracy with no external data. Unlike ViT and DeiT, the~detection transformer (DETR) \cite{carion2020detr} uses both the encoder and decoder parts of the transformer. DETR is a new framework of end-to-end detectors, which attains comparable accuracy and inference speed with Faster R-CNN~\cite{ren2015faster_rcnn}. 

We summarize four ways to use a transformer in STR. (a) Master~\cite{2019Master} uses the decoder of the transformer to predict output sequence. It owns a better training efficiency. In~the training stage, a transformer decoder can predict out all-time steps simultaneously by constructing a triangular mask matrix. (b) a transformer can be used to translate from one language to another. So SATRN~\cite{lee2020satrn} and NRTR~\cite{sheng2019nrtr} adopt the encoder-decoder of the transformer to address the cross-modality between the image input and text output. The~image input represents features extracted by shallow CNN. In~addition, SATRN proposes two new changes in the transformer encoder. It uses an adaptive 2D position encoding and adds convolution in feedforward layer. (c) SRN~\cite{yu2020srn} not only adopts the  transformer encoder to model context but also uses it to reason semantic. (d) the transformer encoder works as a feature extractor including context modeling. Our work uses this method of using the transformer encoder. It is different from recent~methods.

\section{Methodology}
\begin{figure}[ht]

\includegraphics[scale=0.4]{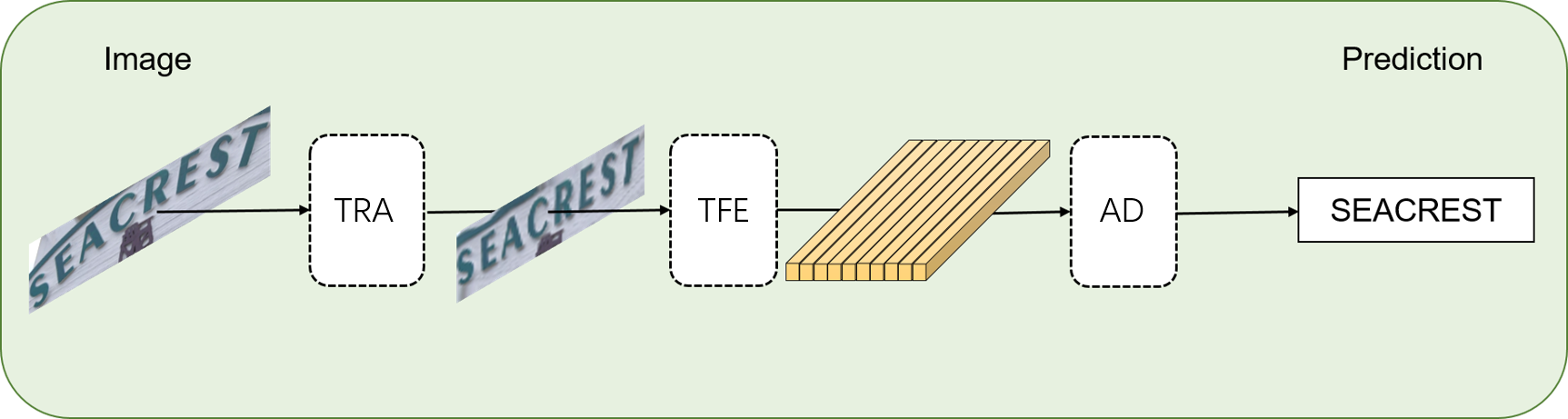}        
\caption{The overview of three-stage text recognizer, TRIG. The~first stage is transformation (TRA), which is used to rectify text images. The~second stage is a Transformer Feature Extractor (TFE), aiming to extract effective and robust features and implicitly model context. The~third stage is an attention decoder (AD), which is used to decode the sequence of features into characters.}
\label{fig_trig}
\end{figure}

This section describes our three-stage text recognition model, TRIG, in~full detail. As~shown in Figure~\ref{fig_trig}, our approach TRIG consists of three stages: Transformation (TRA), Transformer feature extractor (TFE), and~attention decoder (AD). TRA rectifies the input text by a thin-plate spline (TPS) \cite{TPS}. TFE provides robust visual features. AD decodes the feature map to characters. First, we describe the stage of TRA. Second, we show the details of the stage of TFE. Then, the~AD stage is presented. After~that, we introduce the loss function. Finally, we analyze the efficiency of our method with different patch~sizes.

\subsection{Transformation}
Transformation is a stage to rectify the input text images by rectification module. This module uses a shallow CNN to predict several control points, and~then TPS transformation is applied to diverse aspect ratios of text lines. In~this way, the~perspective and curve text can be rectified. Note, the~picture becomes the size of $32\times100$ here.  

\subsection{Transformer Feature~Extractor}

\begin{figure}[ht]

\includegraphics[scale=0.55]{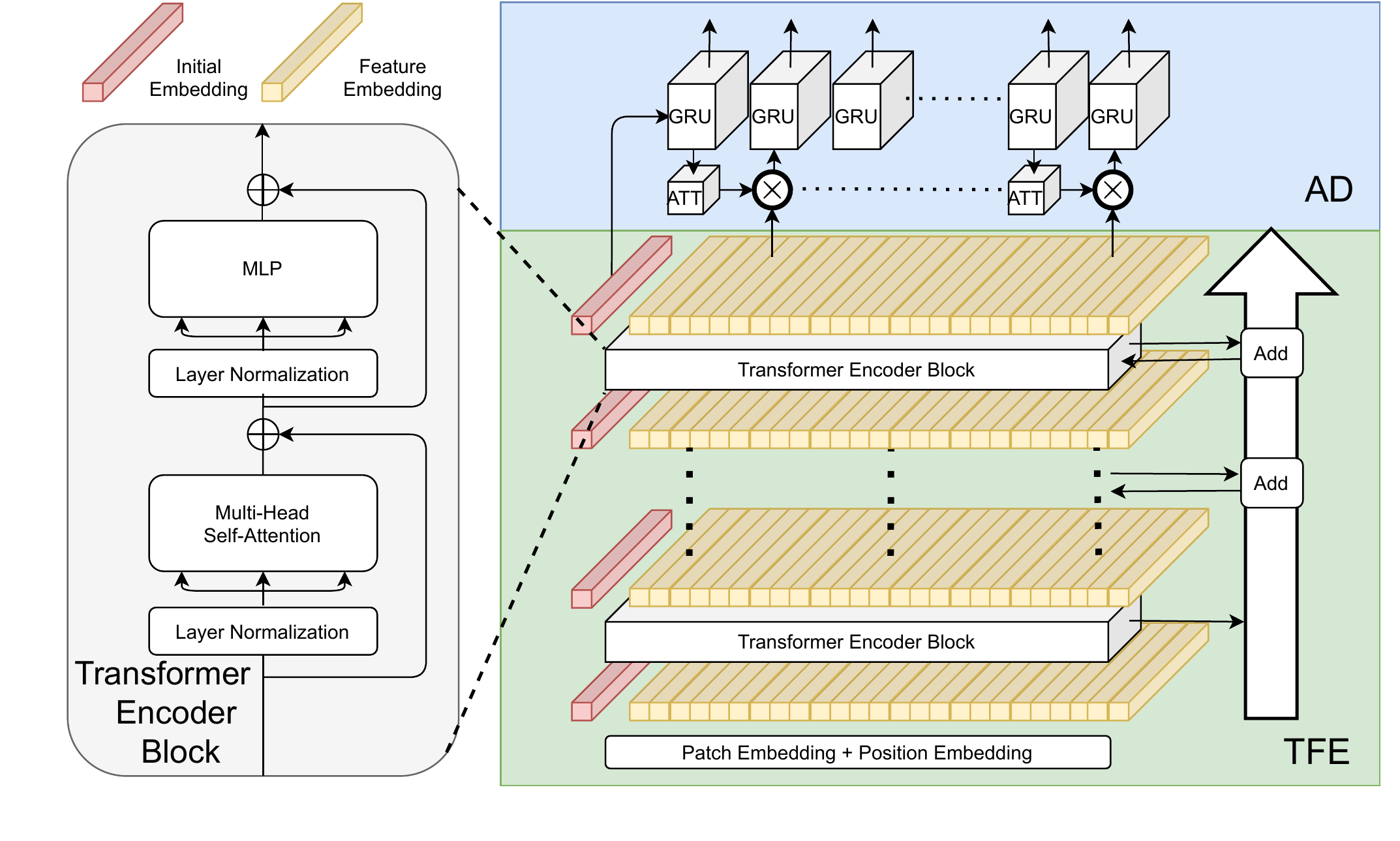}        
\caption{The detailed architectures of the TFE and AD module. TFE consists of Patch Embedding, Position Embedding, Transformer Encoder Blocks, and~Residual add module. The~skip edge in the Residual add module connects Multi-Head Self-Attention modules in adjacent Transformer Encoder Blocks. Here, $\otimes$ denotes matrix multiplication, $\oplus$ denotes broadcast element-wise~addition.}
\label{fig2}
\end{figure}

The TFE is illustrated in Figure \ref{fig2}. In this stage, the transformer encoder is used to extract effective and robust features. First, the~rectified image is split into patches. Unlike the square size of patches in~\cite{2020ViT,2020deit,beal2020vit-frcnn,zhao2020point,valanarasu2021medicaltransformer}, the~rectified image is split into rectangle patches, whose size is $ h\times w $ ($h$ is same as the height of the rectified image). Then the rectified image $ X\in \mathbb{R}^{H\times W\times C} $, where $H, W, C$ is the height, width, and~channel of the rectified image, can be mapped to a sequence $X_s \in \mathbb{R}^{(H \times W \div (h \times w)) \times (3 \times h \times w)}$. Then, a~trainable linear projection $W_E$ {$\in\mathbb{R}^ {(3 \times h \times w) \times D}$ (embedding matrix)} is used to obtain the patch embeddings $E \in \mathbb{R}^{(H \times W \div (h \times w)) \times D}$, where D is the dimension of patch embeddings. The~transcription procedure is given by:
\begin{equation}
E = X_sW_E
\end{equation}

Initial embedding $E_{init}$ is a trainable vector, appended to the sequence of patch embeddings, which goes through transformer encoder blocks, and~is then used to guide the attention decoder. Similar to the role of class token in ViT, we introduce a trainable vector $E_{init}$ called Initial embedding. In~order to encode the position information of each patch embeddings, we use the standard learnable position embeddings. The position embedding can be parameterized by a learnable positional embedding table. For example, position i has i-th position embedding in the embedding table. The position embeddings $E_{pos}$ have the same dimension {$D$} as patch embeddings. At last, the input feature embeddings $F_{0}${$\in \mathbb{R}^{(H \times W \div (h \times w)+1) \times D}$} is the sum of position embeddings and patch embeddings:
\begin{equation}
F_{0} = concat(E_{init}, E) + E_{pos}
\end{equation}

Transformer encoder blocks are applied to the obtained input feature embeddings $F_{0}$. As~we all know, the~transformer encoder block consists of multi-head self-attention (MSA) and multi-layer perceptron (MLP). Following the architecture from ViT, layer normalization (LN) is applied before MSA and MLP. The~MLP contains two linear transformations layers with a GELU non-linearity. The~dimension of input and output is the same, and~the dimension of the inner-layer is four times the output dimension.
The transformer encoder block can be represented as following equations:

\begin{align}
F_l^{'} = MSA(LN(F_{l-1}))+F_{l-1}, &\quad&l = 1,2 \ldots L  \\
F_l = MLP(LN(F_{l}^{'}))+F_l^{'},  &\quad&l = 1,2 \ldots L    
\end{align}

where $l$ denotes the index of blocks, and~$L$ is the index of the last block. {The dimension of $F_l$ and $F_l^{'}$ is $D$ and $4D$.}

To obtain better performance, we add a Residual add module. The~Residual add module uses skip edge to connect MSA modules in adjacent blocks, following Realformer~\cite{he2020realformer}.  
The multi-head process can be unified as:

\begin{align}
MultiHead(Q,K,V,P) = Concat(head_1,\ldots head_h)W^O \\
head_i = Attention(QW_i^Q,KW_i^K,VW_i^V,P) \nonumber \\ 
= Softmax(\frac{QW_i^Q(KW_i^K)^T}{\sqrt{d_k}}+P)VW_i^V .
\end{align}

where query matrix $Q$, key matrix $K$ and value matrix $V$ are linearly projected by $W_i^Q$,  $W_i^K$, $W_i^V$. $P$ is the previous attention score $\frac{QW_i^Q(KW_i^K)^T}{\sqrt{d_k}}+P^{'}$ in the previous~block.

After transformer feature extraction, the~feature map $F_{L} = [f_{init}, f_1,f_2, \ldots f_N ]$ can be obtained. Please note that $f$ means feature embedding. $N$ denotes the index of feature embeddings without $f_{init}$.

\subsection{Attention~Decoder}
The architecture is illustrated in Figure \ref{fig2}. We use an attention decoder to decode the sequence of the feature map. $f_{init}$ is used to initialize the RNN decoder and $[f_1, f_2, \ldots,f_N]$ is used as the input of the decoder. First, we obtain an attention map from the feature map and the internal state from RNN:
\begin{align}
    e_{t,i} = w^Ttanh(W_{d}s_{t-1}+V_{d}f_i+b) \\
    \alpha_{t,i} = \frac{exp(e_{t,i})}{{\sum^{n}_{i^{'}=1}}e_{t,i^{'}}}
\end{align}
where $b, w, W_d, V_d$ are trainable parameters, $s_{t-1}$ is the hidden state of the recurrent cell within the decoder at time $t$. Specifically, $s_0$ is equivalent to $f_\text{init}$. 

Then, we use attention map to element-wise product the feature map $[f_1, f_2, \ldots, f_N]$, and~combine all of them to obtain a vector $g_t$ which called a glimpse:
\begin{align}
    g_t = \sum^{n}_{i=1}\alpha_{t,i}f_i
\end{align}

Next, RNN is used to produce an output vector and a new state vector. The~recurrent cell of the decoder is fed with:
\begin{align}
    (x_t, s_t) = RNN(s_{t-1},[g_t,f(y_{t-1})]),
\end{align}
where $[g_t,f(y_{t-1})]$ denotes the concatenation between $g_t$ and the one-hot embedding of~$y_{t-1}$.

Here, we use GRU~\cite{gru} as our recurrent unit. At~last, the~probability for a given character can be expressed as:
\begin{align}
    p(y_t) = softmax(W_0x_t+b_o).
\end{align}

\subsection{Training~Loss}
Here, we use the standard cross-entropy loss, which can be defined as:
\begin{align}
    L_{CE} = -\sum^{T}_{t=1}(logp(y_t|I)),
\end{align}
where $y_1,y_2,\ldots,y_T$ is the groundtruth text represented by a character~sequence.

\subsection{Efficiency~Analysis}

For the TFE, we assume that the hidden dimension of the TFE is $D$ and that the input sequence length is $T$. The~complexity of self-attention and MLP is $O(T^2\times{D})$ and $O(T\times{D^2})$. Comparing the 1-D split (rectangle patch with the size of 32$\times$4) with the 2-D split (square patch with the size of 4$\times$4), the~sequence lengths are, respectively, 26 and 201, when the input image is 32$\times$100. As~$201^2 \gg 26^2$ and $201>26$, the~complexity and MACs of the 2-D split are far more than the 1-D split. The~MACs of TFE are, respectively, 1.641G and 12.651G, when using 1-D split and 2-D~split.

For the decoder, the~complexity gap between 1 and D split and 2-D split comes from the process of obtaining an attention map and glimpse. We assume that the sequence length is $T$. The~complexities are both $O(T)$. Therefore, the shorter sequence has lower complexity. The~MACs of AD using the 1-D split is 0.925G when it is 5.521G using the 2-D~split.    

Based on the above analysis, we propose to use a 1-D split to increase~efficiency.

\section{Experiments}

In this section, we demonstrate the effectiveness of our proposed method. First, we give a brief introduction to the datasets and the implementation details. Then, our method is compared with state-of-the-art methods on several public benchmark datasets. Next, we make some discussions on our method. Finally, we perform ablation studies to analyze the performance of different~settings.  

\subsection{Dataset}
In this paper, models are only trained on two public synthetic datasets MJSynth (MJ) \cite{jaderberg2014mj} and SynText (ST) \cite{gupta2016synthetic} without any additional synthetic dataset, real dataset or data augmentation. There are 7 scene text benchmarks chosen to evaluate our~models.

MJ contains 9 million word box images, which is generated from a lexicon of 90K English~words.

ST is a synthetic text dataset generate by an engine proposed in~\cite{gupta2016synthetic}. We obtain 7 million text lines from this dataset for~training. 

IIIT5K (IIIT) \cite{mishra2012iiit} contains scene texts and born-digital images, which are collected from the website. It consists of 3000 images for~evaluation.

Street View Text (SVT) \cite{wang2011svt} is collected from the Google Street View. It contains 647~images for evaluation, some of which are severely corrupted by noise and blur or have low~resolution.

ICDAR 2003 (IC03) \cite{lucas2005ic03} have two different versions of the dataset for evaluation: versions with 860 or 867 images. We use 867 images for~evaluation. 

ICDAR 2013 (IC13) \cite{ic13} contains 1015 images for~evaluation.

ICDAR 2015 (IC15) \cite{ic15} contains 2077 images, captured by Google Glasses, some of which are noisy, blurry, and~rotated or have low resolution. Researchers have used two different versions for evaluation: 1811 and 2077 images. We use both of~them.

SVT-Perspective (SVTP) \cite{svtp} 
contains 645 cropped images for evaluation. Many of the images contain perspective projections due to the prevalence of non-frontal~viewpoints.

CUTE80 (CUTE) \cite{cute} contains 288 cropped images for evaluation. Many of these are curved text~images.

\subsection{Implementation~Details}
The proposed TRIG is implemented in the PyTorch framework and trained on two RTX2080Ti GPUs with 11GB memory. As~for the training details, we do not perform any type of pre-training. The~decoder recognizes 97 character classes, including digits, upper-case and lower-case letters, 32 ASCII punctuation marks, end-of-sequence (eos) symbol, padding symbol, and~unknown symbol. We adopt the AdaDelta optimizer and the decayed learning rate. Our model is trained on ST and MJ for 24 epochs with a batch size of 640, the~learning rate is set to 1.0 initially and decayed to 0.1 and 0.01 at the 19th and the 23rd epoch. The~batch is sampled 50\% from ST and 50\% from MJ. All images are resized to $64\times256$ during both training and testing. We use the IIIT dataset to select our best~model.

At inference, we resize the images to the same size as for training. Furthermore, we use beam search by maintaining five candidates with the top accumulative scores at every step to~decode.  

\subsection{Comparison with~State-of-the-Art}
We compare our methods with previous state-of-the-art methods on several benchmarks. The~results are shown in Table~\ref{tab1}. Even compared to reported results using additional real or private data and data augmentation, we achieve satisfying performance. Compared with other methods trained only by MJ and ST, our method achieves the best results on four datasets including SVT, IC13, IC15, and~SVTP. TRIG provides an accuracy of +1.5~percentage point (pp, 93.8\% vs. 92.3\%) on SVT, +0.2 pp (95.2\% vs. 95.0\%) on IC13, +1.6 pp (81.1\% vs. 79.5\%) on IC15 and +1.6 pp (88.1\% vs. 86.5\%) on SVTP. 

\begin{table}
\caption{Comparison with SOTA methods. 'SA', 'R' indicates using SynthAdd dataset or real datasets apart from MJ and ST. 'A' means using data augmentation when training. For a fair comparison, reported results using additional data or data augmentation are not taken into account. Top accuracy for each benchmark is shown in \textbf{bold}. The average accuracy is calculated on seven datasets and IC15 using the version which contains 2077 images.}\label{tab1}
\scalebox{0.9}{
\begin{tabular} {ccc c c c c c c c c}
\hline
\multirow{2}*{Method} & \multirow{2}*{Training data} &  IIIT & SVT & IC03 & IC13 & IC15 & IC15 & SVTP & CUTE & \multirow{2}*{Average}\\
& &  3000   & 647 & 867  & 1015 & 1811 & 2077 & 645  & 288 & \\
\hline
CRNN \cite{shi2016crnn} & MJ &  78.2 & 80.8 & - & 86.7 & -  & -  & - & - & -\\
FAN \cite{cheng2017fan}  & MJ+ST & 87.4 & 85.9 & 94.2 & 93.3 & - & 70.6 & - & - & -  \\
ASTER \cite{2018ASTER} & MJ+ST & 93.4 & 89.5 & 94.5 & 91.8 & 76.1 & - & 78.5 & 79.5 & -\\
SAR \cite{li2019sar}  & MJ+ST & 91.5 & 84.5 & - & 91.0 & - & 69.2 & 76.4 & 83.3 & -\\
ESIR \cite{zhan2019esir} & MJ+ST & 93.3 & 90.2 & - & 91.3  & 76.9 & - & 79.6 & 83.3 & -\\
MORAN \cite{luo2019moran}  & MJ+ST & 91.2 & 88.3 & 95.0 & 92.4 & - & 68.8 & 76.1 & 77.4 & 84.5\\
SATRN \cite{lee2020satrn} & MJ+ST & 92.8 & 91.3 & \textbf{96.7} & 94.1 & - & 79 & 86.5 & 87.8 & 89.2\\
DAN \cite{wang2020dan} &MJ+ST & 94.3 & 89.2 & 95.0 & 93.9 & - & 74.5 & 80.0 & 84.4 & 87.7\\
RobustScanner \cite{yue2020robustscanner} & MJ+ST &  \textbf{95.3} & 88.1 & - & 94.8 & - & 77.1 & 79.2 & \textbf{90.3} & -\\
PlugNet \cite{plugnet} & MJ+ST & 94.4 & 92.3 & 95.7 & 95.0 & 82.2 & - & 84.3 & 85.0 & - \\
AutoSTR \cite{zhang2020autostr} & MJ+ST &{94.7}  & {90.9}  & 93.3  & {94.2}  & {81.8} & -  & {81.7}  & -  & - \\
GA-SPIN \cite{zhang2020spin} & MJ+ST &95.2  &90.9  & -  & 94.8  &  82.8 & 79.5 &  83.2  &  {87.5} & -\\
\hline
MASTER \cite{2019Master} & MJ+ST+R & 95 & 90.6 & 96.4 & 95.3 & - & 79.4 & 84.5 & 87.5 & 90.0\\
GTC \cite{hu2020gtc} & MJ+ST+SA & 95.5 & 92.9 & - & 94.3& 82.5 & - & 86.2 & {92.3} & -\\ 
SCATTER \cite{litman2020scatter} & MJ+ST+SA+A & 93.9 & 90.1 & 96.6 & 94.7 & - & 82.8 & 87 & 87.5 & 90.5\\
RobustScanner \cite{yue2020robustscanner} & MJ+ST+R &{95.4} & 89.3  &  - & 94.1 & -  & 79.2  & 82.9 & 92.4  & -\\
SRN \cite{yu2020srn} & MJ+ST+A &  94.8 & 91.5 & - & 95.5 & 82.7 & - & 85.1 & 87.8 & -\\
\hline
TRIG & MJ+ST & 95.1 & \textbf{93.8} & 95.3 & \textbf{95.2} & \textbf{84.8} & \textbf{81.1} & \textbf{88.1} & 85.1 & \textbf{90.8}\\
\hline
\end{tabular}
}
\end{table}

\subsubsection{Discussion on Training~Procedures}
\begin{figure}[ht]
\centering
\includegraphics[scale=0.5]{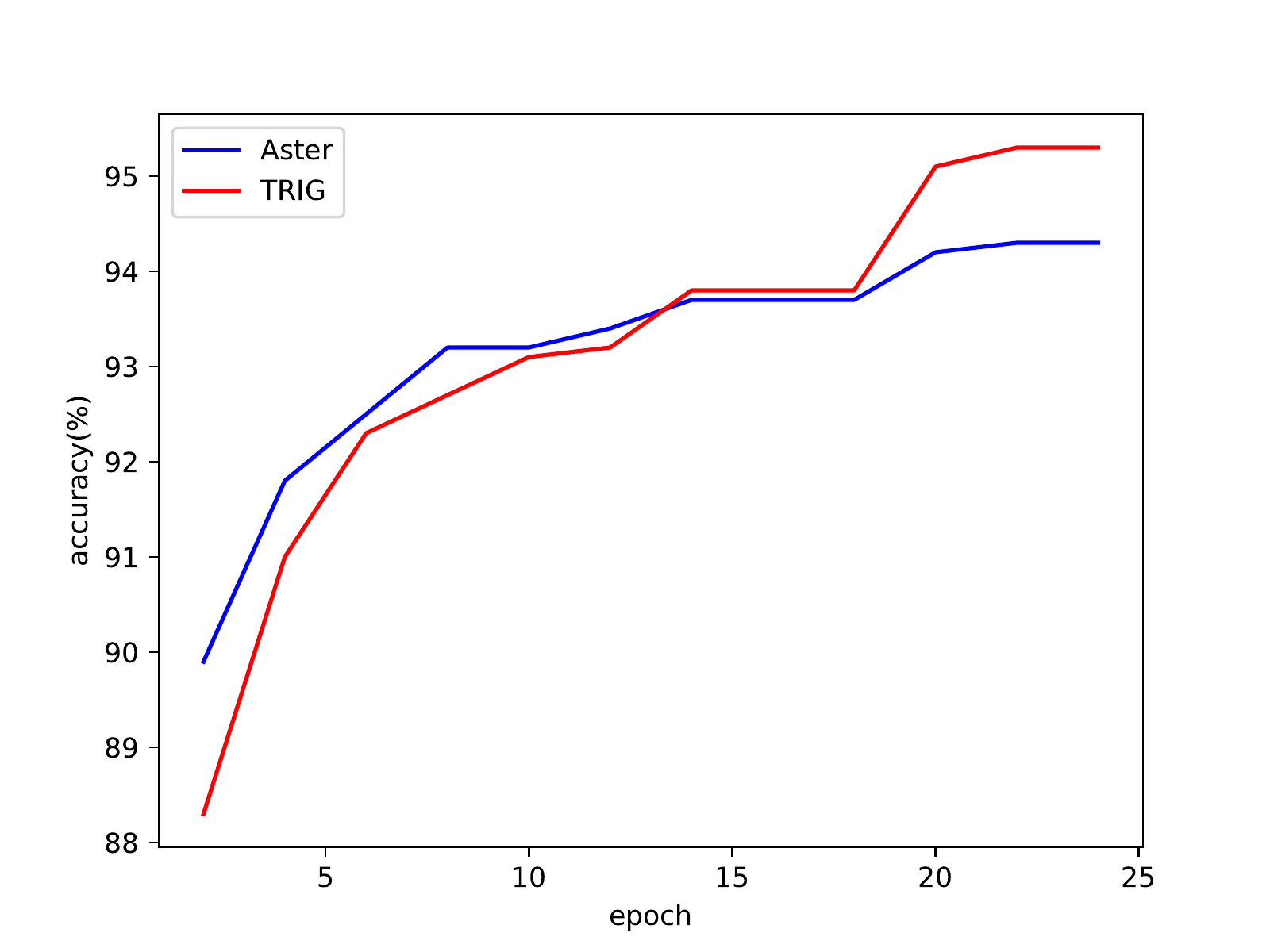}        
\caption{The accuracy on IIIT between TRIG (Ours) and ASTER during the training~process.}
\label{fig_epoch}
\end{figure}

Our method TRIG needs long training epochs. We train TRIG and ASTER with the same learning rate and optimizer for 24 epochs. The~whole training procedure takes a week. The~learning rate, which is 1.0, drops by a factor of 10 after 19 and 23 epochs. As~shown in Figure~\ref{fig_epoch}, the~accuracy of TRIG on IIIT is lower than ASTER in the preceding epoch and better than ASTER after 14 epochs. At~last, the~accuracy of TRIG on IIIT and its average accuracy on all datasets is 0.8pp and 2pp higher than ASTER. We can conclude that the transformer feature extractor needs longer epochs to train to improve the accuracy. Because~the transformer does not have properties such as CNN (i.e., shift, scale, and~distortion~invariance).

\subsubsection{Discussion on Long-Range~Dependencies}
\begin{figure}[ht]
\centering
\includegraphics[scale=0.3]{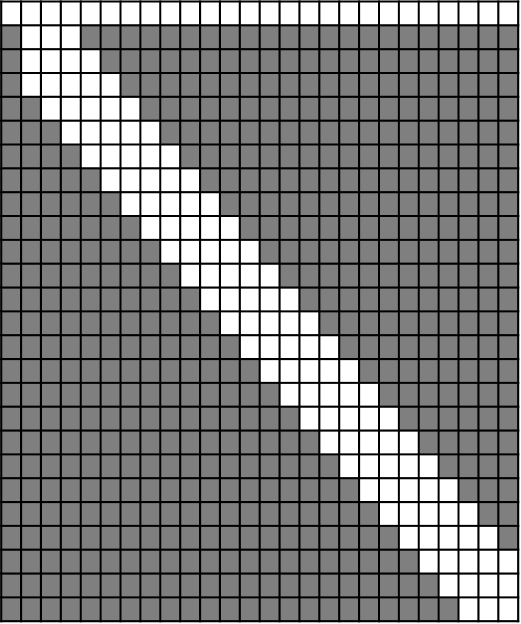}       
\caption{The detail of the mask {used in the transformer encoder}. The~gray square denotes the mask. The~first row means initial embedding which can see all embeddings. The~remaining features can only see other features of the window~size.}
\label{fig_mask}
\end{figure}

The decoder of the transformer uses triangle masks to promise that the prediction of one-time step $t$ can only access the output information of its previous steps. Taking inspiration from this, we use the mask to let the transformer encoder only access the nearby feature embeddings. By~this, the transformer encoder will have a small reception. We empirically analyze the performance of long-range dependencies by comparing two kinds of methods with or without masks, demonstrated in Table~\ref{tab2}. The~detail of the mask is shown in \mbox{Figure~\ref{fig_mask}}. We observe the decrease of accuracy when using mask no matter CTC decoder and attention decoder. It indicates that long-range dependencies are related to effectiveness and the feature extractor of our method can capture long-range~dependencies.

\begin{table}
\begin{center}
\caption{Performance comparison with mask and context modeling module. There are two kinds of decoder, CTC and attention decoder. All models are trained on two synthetic datasets. The batch size is 896 and the training step is 300000. The number of blocks in TFE is 6 and skip attention is not used.}\label{tab2}
\scalebox{1}{
\begin{tabular}{cccc c c c c c c}
\hline
\multirow{2}*{Mask} & Context &  \multirow{2}*{Decoder} & IIIT & SVT & IC03 & IC13  & IC15 & SVTP & CUTE\\
&modeling&  & 3000   & 647 & 867  & 1015 & 2077 & 645  & 288\\
\hline
$\times$     & $\checkmark$ & CTC & 85.0 & 80.4 & 91.7 & 89.6 &65.8 &70.5 & 69.4 \\
$\checkmark$ & $\times$     & CTC &85.6 & 80.5 & 90.9 & 88.4  & 63.1 & 65.6 & 67.7 \\
$\times$     & $\times$     & CTC & 86.5 & 82.1 & 92.0 & 89.6  & 65.9 & 71.9 & 71.2 \\
\hline
$\times$     & $\checkmark$ & Attn & 87.9 & 86.7 & 92.3 & 90.8  & 71.9 & 79.4 & 72.9\\
$\checkmark$ & $\times$     & Attn & 86.7 & 83.5 & 91.6 & 89.5 & 70.7 & 78.5 & 67.7\\
$\times$     & $\times$     & Attn & 89.5 & 88.6 & 93.7 & 92.5& 74.1 & 79.2 & 76.4\\
\hline
\end{tabular}}
\end{center}
\end{table}

\subsubsection{Discussion on Extra Context Modeling~Module}
As the effective module of context modeling is used in STR networks, we consider whether we need an extra context modeling module when using the transformer encoder to extract features. After~the transformer feature extractor, a~context modeling module (BiLSTM) is added. As~shown in Table~\ref{tab2}, the accuracy of almost all data sets is lower than methods without BiLSTM (except +0.2pp on SVTP); therefore, we conclude that it is not necessary to add BiLSTM after the transformer feature extractor to the model context, because~the transformer feature extractor has an implicit ability to model context. An extra context modeling module may break the context modeling that the model has implicitly~learned.  

\subsubsection{Discussion on Patch~Size}
Table~\ref{tab_patch} shows the average accuracy in seven scene text benchmarks with different patch sizes when the models are trained with the same settings. The~number of blocks in TFE is 6 and the batch size is 192. The~training step of all models is 300,000 and skip attention is not used. The~transformation stage is used for 1-D split because it loses the information on height dimension. The~rest of the methods do not use a transformation stage, because~the feature map is 2-D and a 2-D attention decoder can be used for it to have a glimpse of the character feature. When using 1-D split, the~method with the patch size of $32\times4$ is better than other patch sizes. Therefore, we set this patch size as the patch size of TRIG. Furthermore, our method with 1-D split (patch size of 32~$\times~4$) achieves better average accuracy than 2-D split (patch size of 16~$\times~$4, 8~$\times$~4, and~4~$\times$~4). Therefore, the~design of 1-D split of rectangle patch with the transformation stage can work better than the other patch size with the 2-D attention~decoder.   
\begin{table}[h]
\centering
\caption{Performance comparison with different patch size. 'Average' denotes the average accuracy of seven standard benchmarks. * indicates that the model did not converge, possibly because sequence length is too short to decode.}\label{tab_patch}
\begin{tabular}{cccc}
\hline
 & Patch Size                 & Sequence length            & Average \\ \hline
 1-D split & 32 $\times$ 2 & 1 $\times$ 50 & 80.9   \\ \hline
 1-D split & 32 $\times$ 4 & 1 $\times$ 25 & \textbf{83.5}   \\ \hline
 1-D split & 32 $\times$ 5 & 1 $\times$ 20 & 81.0   \\ \hline
 1-D split & 32 $\times$ 10 & 1 $\times$ 10 & 81.7   \\ \hline
 1-D split & 32 $\times$ 20 & 1 $\times$ 5 & $0.1^{*}$  \\ \hline
 2-D split & 16 $\times$ 4 & 2 $\times$ 25 & 81.9       \\ \hline
 2-D split & 8 $\times$ 4  & 4 $\times$ 25 & 73.5   \\ \hline
 2-D split & 4  $\times$ 4 & 8 $\times$ 25 & 76.7   \\ \hline
\end{tabular}
\end{table}

\subsubsection{Discussion on Initial~Embedding}
\begin{figure}[ht]
\centering
\includegraphics[scale=0.46]{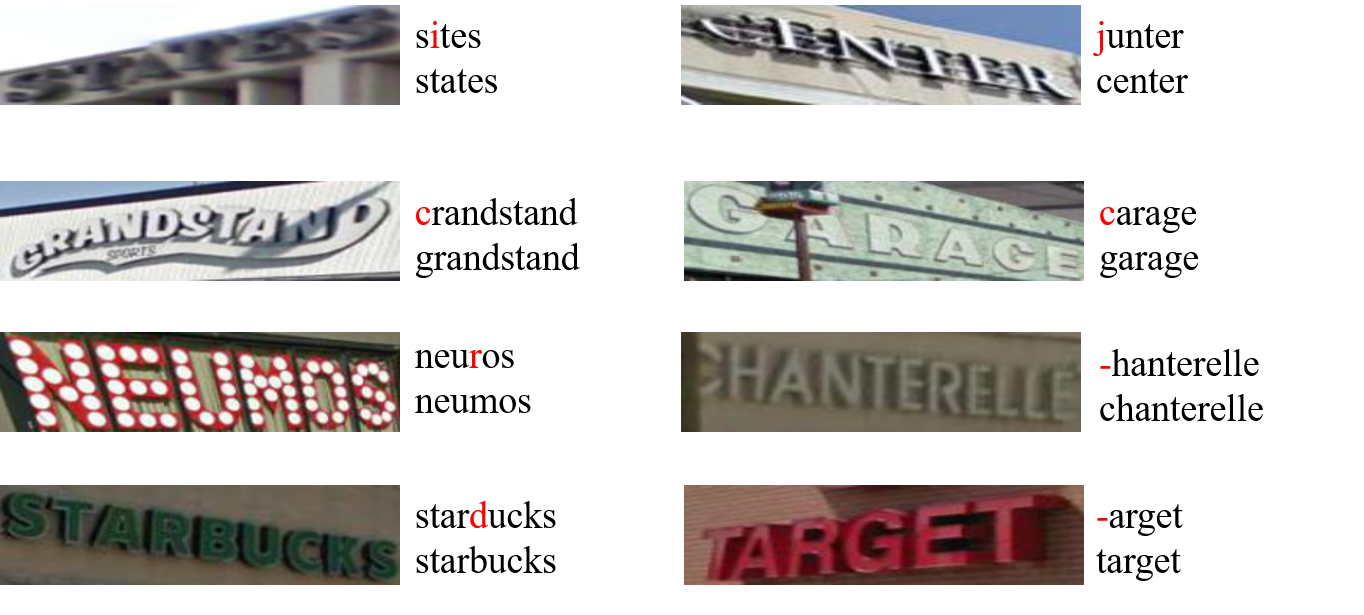}      
\caption{Right cases of TRIG with/without initial embedding guided. The~predictions are placed along the right side of the images. The~top string is the prediction of TRIG without the initial embedding guidance. The~bottom string is the prediction of~TRIG.}
\label{fig4}
\end{figure}

To illustrate how the learnable initial embedding helps TRIG improve performance, we collect some individual cases from the benchmark datasets to compare the predictions of TRIG with and without the learnable initial embedding. As~shown in Figure~\ref{fig4}, the~prediction without initial embedding guided may lose the first character or mispredict the first character while TRIG with initial embedding predicts the right character. Furthermore, the~initial embedding can also bring the benefit to decode all characters, such as 'starducks' to~'starbucks'.

\subsubsection{Discussion on Attention~Visualization}
\begin{figure}[ht]
\includegraphics[scale=0.4]{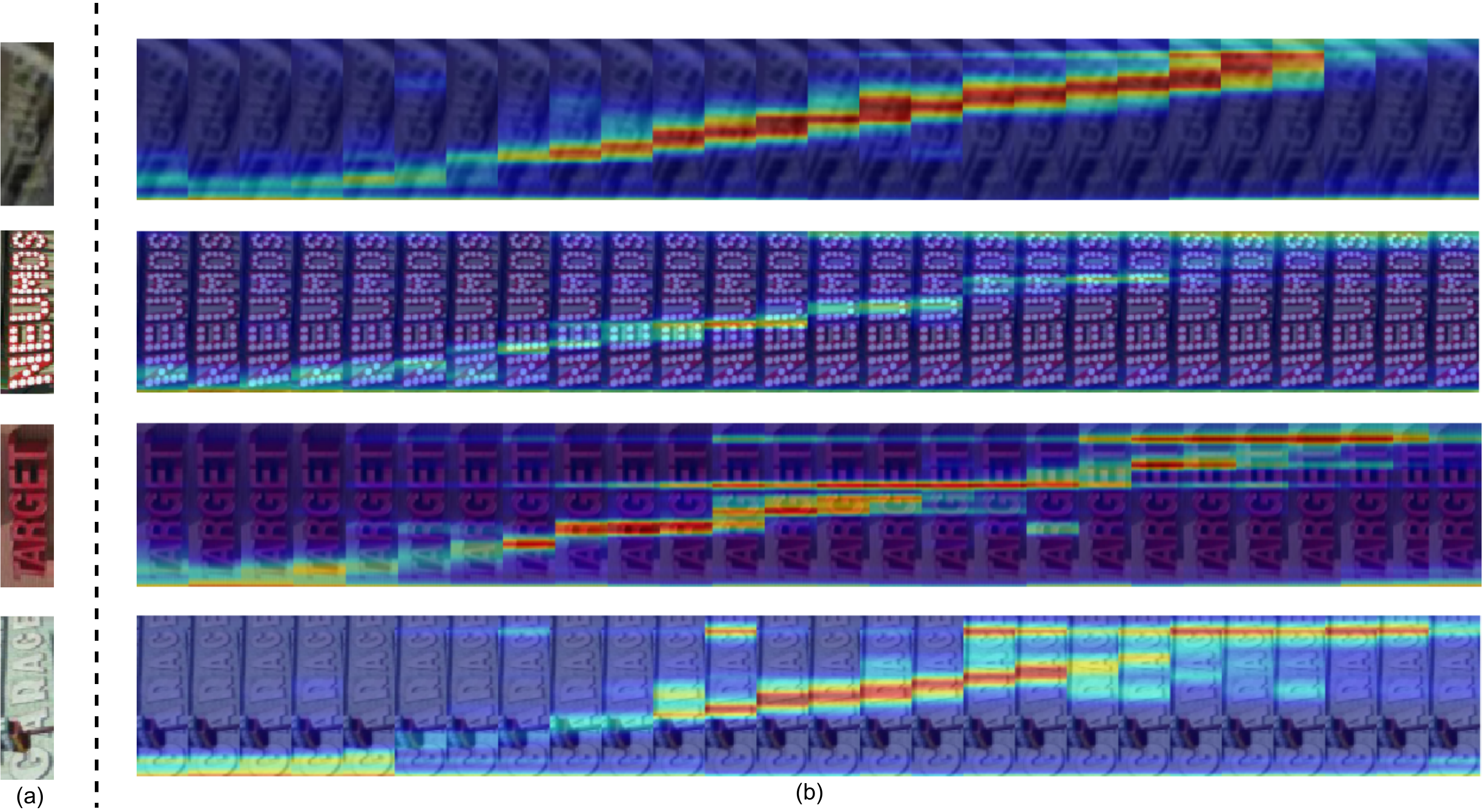}       
\caption{{The attention visualization of TFE.} (\textbf{a}) Examples of rectified images. (\textbf{b}) The attention map of initial embedding and each embedding in feature map extracted with the transformer~encoder. }
\label{fig3}
\end{figure}

Figure~\ref{fig3} shows the attention map from each embedding to the input images. We used Attention Rollout~\cite{abnar2020attention_roll_out} following ViT. We averaged the attention weights of TRIG across all heads and recursively compute the attention matrix. At~last, we can get pixel attribution for embeddings in (b). The~first row shows what part of the rectified picture is responsible for the initial embedding. Furthermore, the rest of the rows represent the attention map of feature embeddings $[f_1, f_2, \ldots, f_N]$. The~initial embedding is mainly relevant to the first character. For~each feature embedding, some of the features come from adjacent embeddings and others come from distant embeddings. According to Figure~\ref{fig3}, we can roughly learn how the transformer feature extractor extract features and model long-range~dependencies.  

\subsubsection{Discussion on~Efficiency}
To verify the efficiency of our model, we compare the MACs, parameters, GPU memory, and~inference speed of our model using 1-D split (square patch with the size of 4~$\times$~4) and 2-D split (rectangle patch with the size of 32~$\times$~4) and ASTER. The~result is shown in Table~\ref{tab_speed}. The~MACs of TRIG with 1-D split and training GPU memory cost (the batch size is 32) are 7 times and 3.4 times of the method with 2-D split. Furthermore, the~speed of TRIG is faster than ASTER. The~speed performance is tested on GeForce GTX 1080 Ti~GPU.

\begin{table}[ht]
\caption{Efficiency comparison between ASTER and TRIG. MACS, model parameters, gpu memory cost, and speed are compared.}\label{tab_speed}
\centering
\scalebox{1}{
\begin{tabular}{cccccc}
\hline
Method     & \begin{tabular}[c]{@{}c@{}}MACs\\  G\end{tabular} & \begin{tabular}[c]{@{}c@{}}\#param.\\     M \end{tabular} & \begin{tabular}[c]{@{}c@{}}GPU memory\\ (train)  m\end{tabular} & \begin{tabular}[c]{@{}c@{}}GPU memory\\ (inference)  m\end{tabular} & \begin{tabular}[c]{@{}c@{}}inference time\\     ms/image\end{tabular} \\ \hline
ASTER      & 1.6                                               & 21.0                                                           & 1509                                                            & 3593                                                                & 19.5                                                                  \\ \hline
TRIG  & 2.6                                               & 68.1                                                           & 2579                                                            & 1855                                                                & 16.2                                                                  \\ \hline
2-D split & 18.2                                              & 68.0                                                           & 8717                                                            & 1929                                                                & 37.6                                                                  \\ \hline
\end{tabular}
}
\end{table}

\subsection{Ablation~Study}

In this section, we perform a series of experiments to evaluate the impact of blocks, heads, embedding dimension,  initial embedding guide, and~skip attention on recognition performance. All models are trained from scratch on two synthetic datasets. The~results are reported on seven standard benchmarks and shown in Tabel \ref{tab_ablation}. We can make the following observations: (1) The TRIG with 12 blocks is better than the model of six blocks. The~performance can be improved by stacking more transformer encoder blocks. However, when transformer encoder goes deeper, the~stacking blocks cannot increase the performance (the average accuracy of 24 blocks decreases by 0.2pp). The~accuracy may reach the bottleneck. It is expected that stacking blocks lead to more challenging training procedures. (2) Initial embedding can bring gains to the model on average accuracy no matter skip attention is applied or not. This shows the effectiveness of initial embedding guidance. (3)~Skip attention is important to accuracy. Regardless of the depth of the feature extractor, the~addition of skip attention brings gains to the performance. (4) When other conditions are guaranteed to be the same, the~average accuracy of 16 heads is better than the condition of 8 heads. Besides, the~average accuracy of 512 dimensions is better than 256~dimensions. 

\begin{table}[ht]
\begin{center}
\caption{Performance comparison with different settings. 'Initial guidance' indicates the initial embedding guidance. 'Skip attention' denotes the skip attention in the TFE. 'Average' means the average accuracy of seven standard benchmarks. * denotes the setting of TRIG shown in Table \ref{tab1}.}\label{tab_ablation}
\begin{tabular}{|ccccc|c|}
\hline
\multirow{2}*{Blocks} & \multirow{2}*{Heads} & Embedding  & Initial  &Skip   & \multirow{2}*{Average}\\
        &       & dimension  & guidance    &attention  & \\
\hline
6 & 16 & 512 & $\times$     & $\times$     & 89.2\\  
6 & 16 & 512 & $\times$     & $\checkmark$ & 89.8(+0.6)\\
12& 16 & 256 & $\times$     & $\times$     & 89.4(+0.2)\\
12& 8  & 256 & $\times$     & $\checkmark$ & 90.1(+0.9)\\
12& 16 & 256 & $\times$     & $\checkmark$ & 90.4(+1.2)\\
\hline
12& 16 & 512 & $\times$     & $\times$     & 89.5(+0.3)\\
12& 16 & 512 & $\checkmark$ & $\times$     & 89.6(+0.4)\\
12& 16 & 512 & $\times$     & $\checkmark$ & 90.6(+1.4)\\
12& 16 & 512 & $\checkmark$ & $\checkmark$ & \textbf{90.8}(\textbf{+1.6})*\\
\hline
24& 16& 512 & $\checkmark$ & $\checkmark$  & 90.6(1.4)\\
\hline
\end{tabular}
\end{center}
\end{table} 

\section{Conclusions}
In this work, we propose a three-stage transformer-based text recognizer with initial embedding guidance named TRIG. In~contrast to the existing STR network, this method only uses transformer feature extractor to extract robust features and does not need a context modeling module. A~1-D split is designed to divide text images. Besides, we propose a learnable initial embedding learned from transformer encoder to guide the attention decoder. Extensive experiments demonstrate that our method sets the new state of the art on several~benchmarks. 

We also demonstrate that the longer training epochs and long-range dependencies are essential to~TRIG.

We consider two promising directions for future work. First, a~better transformer architecture that is more suitable for Scene text recognition can be designed to extract more robust features. For~example, a transformer can be designed as a pyramid structure such as CNN or some other structure. Second, we see potential in using a transformer in an end-to-end text spotting system.

%
%
%
\bibliographystyle{splncs04}
\bibliography{reference}
%





\end{document}